\documentclass{article}
\usepackage{amsfonts}
\usepackage{iclr2026_conference,times}

\usepackage{amsmath,amsfonts,bm}

\def\eqref#1{equation~\ref{#1}}

\def\1{\bm{1}}

\DeclareMathAlphabet{\mathsfit}{\encodingdefault}{\sfdefault}{m}{sl}
\SetMathAlphabet{\mathsfit}{bold}{\encodingdefault}{\sfdefault}{bx}{n}

\usepackage{hyperref}
\usepackage{url}
\usepackage{graphicx}
\usepackage{subcaption}
\usepackage{amssymb}

\title{When Quotes Crumble: Detecting Transient Mechanical Liquidity Erosion in Limit Order Books}

\author{%
\textbf{Haohan Xu}$^{1}$\thanks{Main contributor},
\textbf{Jason Bohne}$^{1,2}$,
\textbf{Pawe{\l} Polak}$^{1,2}$\thanks{Corresponding author},\\
\textbf{Yurij Baransky}$^{2}$,
\textbf{Ajay Alva}$^{2}$,
\textbf{Violetta Fedotova}$^{2}$,
\textbf{Gary Kazantsev}$^{2}$,
\textbf{David Rosenberg}$^{2}$ \\
$^1$Stony Brook University \quad
$^2$Bloomberg\\
\texttt{\{haohan.xu, jason.bohne, pawel.polak\}@stonybrook.edu, \{ybaransky,} \\
\texttt{aalva, vfedotova, gkazantsev, drosenberg44\}@bloomberg.net}
}

\iclrfinalcopy 
\begin{document}

\maketitle

\begin{abstract}

We study the detection of transient liquidity erosion (``crumbling quotes") in electronic limit order books, where observable quote deterioration may reflect either mechanical liquidity withdrawal or informational repricing. Using the ABIDES agent-based simulator, we construct a multi-agent environment in which crumbling emerges from stochastic regime switches in a market maker, providing time-resolved ground truth unavailable in real market data. We develop a detection pipeline that identifies mechanically driven quote erosion using order book features, and train a neural model to produce calibrated crumbling probabilities. Experiments demonstrate that the proposed framework reliably identifies crumbling events against agent-level ground truth, with the neural model achieving +36\% AUC improvement over rule-based baselines and robust performance across normal, high-volatility, bull, and bear market conditions. Ablation studies on temporal features and varying the dependence structure of the ground-truth mechanism confirm that the framework generalizes across both independent and autocorrelated liquidity withdrawal dynamics.

\end{abstract}

\begin{figure}[htbp]
\begin{center}
\includegraphics[width=1.0\textwidth]{iclr2026/NewFigure1.png}
\end{center}
\caption{
\textbf{Distinguishing transient mechanical quote erosion from persistent repricing.}
\textbf{Left:} Short-horizon limit order book deterioration is observationally ambiguous, arising from either transient mechanical liquidity withdrawal (crumbling) or persistent information-driven repricing.
\textbf{Middle:} An agent-based market simulation induces controlled, finite-duration liquidity-withdrawal regimes, providing time-resolved, side-specific ground truth unavailable in real markets.
\textbf{Right:} A mechanics-first detection pipeline leverages transience and microstructural constraints to produce interpretable, probabilistic labels of transient quote erosion from LOB data.
}
\label{fig:architecture}
\end{figure}

\section{Introduction and Related Work}
\label{sec:intro}

Learning-based trading and execution systems operate in multi-agent markets where displayed liquidity can change rapidly.
A common execution failure occurs when visible depth at the best quote is removed faster than it is replenished, causing the best bid or ask to deteriorate by multiple ticks.
Following standard market microstructure terminology, we refer to such episodes as \emph{crumbling quotes}~\citep{cartea2015algorithmic}.
These events widen spreads and increase execution costs when decisions are based on quotes that no longer reflect immediately available liquidity.

Detecting crumbling quotes from limit order book (LOB) data is challenging because observed quote deterioration is not uniquely attributable to liquidity withdrawal~\citep{zhang2019deeplob, kolm2023deep}.
As illustrated in the left panel of Figure~\ref{fig:architecture}, short-horizon quote deterioration may arise either from \emph{mechanical quote erosion}, where displayed depth is removed while latent value remains stable, or from \emph{informational repricing}, where quotes adjust to new information.
Although both mechanisms produce similar top-of-book signatures, their implications differ: mechanical erosion is typically transient, whereas informational repricing reflects a persistent change in the efficient price.
This observational ambiguity complicates both offline evaluation and online control of execution policies.
Classical optimal execution models~\citep{bertsimas1998optimal,almgren2001optimal,foucault2005limit} and benchmark schedules such as TWAP, VWAP, and implementation shortfall~\citep{kissell2013science} do not explicitly account for transient liquidity withdrawal.
Recent reinforcement learning approaches to execution~\citep{nevmyvaka2006reinforcement,buehler2019deep,beysolow2019market,ning2021double,theate2021application,xu2025learning} similarly depend on reliable identification of adverse microstructure regimes during training and evaluation.

Crumbling quotes are closely linked to adverse selection in liquidity provision~\citep{glosten1985bid,kyle1985continuous,easley2012flow}.
When fast participants react to price movements, displayed liquidity may be consumed before slower agents can cancel or reprice, exposing liquidity providers to losses and generating transient quote instability~\citep{brogaard2014high}.
Recent theoretical work models such order-flow dynamics with Hawkes processes, capturing self-excitation and its links to market impact and volatility.
Empirical and theoretical studies of latency advantages and speed races motivate this mechanism-level view~\citep{hendershott2011does,shkilko2020every,aquilina2022quantifying}.
At the market design level, continuous limit order books have been argued to incentivize latency arbitrage and adverse selection~\citep{budish2015high,kyle2016market}, motivating interventions such as speed bumps and protective order types~\citep{baldauf2020high}.
Across these settings, a common measurement challenge remains: to distinguish transient, mechanically induced liquidity erosion from value-driven repricing using LOB data alone.

A central challenge is the absence of ground truth in historical market data.
LOB feeds expose messages and aggregated depth but do not reveal whether quote movements arise from liquidity withdrawal or information arrival.
As a result, observational labeling rules may incur false positives and false negatives, with error rates that are difficult to quantify or audit.

\noindent\textbf{Goal and approach.}
The goal of this paper is to construct LOB-only detection and labeling rules for crumbling quotes whose errors can be systematically audited in a controlled environment with known ground truth.
\emph{Importantly, this work does not examine the market forces that induce crumbling quotes.}
Rather, we focus on the \emph{measurement and labeling problem}: given LOB observations, how can crumbling episodes be identified and assigned calibrated labels in a mechanically interpretable, mechanism-agnostic, and auditable manner.
Figure~\ref{fig:architecture} provides a schematic overview of this problem formulation and our solution approach.

To this end, we implement our approach in ABIDES~\citep{byrd2019abides,byrd2020abides}, an agent-based market simulator with configurable latencies and NASDAQ-style messaging.
As shown in the middle panel of Figure~\ref{fig:architecture}, we augment an adaptive market maker with regime-switching mechanisms that vary a quote-skewness parameter $\beta$, inducing finite-duration, side-specific liquidity withdrawal.
Because the regime state is directly observable from the agent policy, this construction yields time-resolved ground-truth intervals for mechanically induced liquidity erosion that are unavailable in real market data.
We further consider alternative agent-level crumbling dynamics with different temporal dependence structures to assess robustness.

\noindent\textbf{Mechanics-constrained detection and calibrated labeling.}
Building on this ground truth, we develop a mechanics-first, LOB-only detection pipeline that identifies candidate events via visible-depletion conditions and excludes informational repricing using efficient-price stability, opposite-side stability, and transience filters.
As summarized in the right panel of Figure~\ref{fig:architecture}, these hard constraints restrict labeling to mechanically interpretable events and enable systematic inspection of failure modes.
On top of this construction, we introduce a neural continuous labeling model that maps event-level features to calibrated crumbling probabilities, trained against agent-level ground truth with a KL (cross-entropy) objective while preserving the hard mechanical eligibility gate.

\noindent\textbf{Contributions}
This paper makes three contributions:
\begin{enumerate}
    \item \textbf{Endogenous ground truth for crumbling in ABIDES:} a controlled, agent-level construction that yields mechanically valid liquidity-withdrawal episodes with known timing and side.
   \item \textbf{LOB-only, mechanics-constrained detection:} a mechanics-first event construction
and feature set for detecting transient mechanical liquidity erosion, combining visible depletion,
efficient-price stability, one-sidedness, transience, and severity--recovery features, defined
purely from observable LOB data.
    \item \textbf{Continuous probabilistic labeling via KL calibration:} a neural labeling model that produces calibrated crumbling probabilities under a hard mechanical gate, enabling severity ranking and uncertainty-aware downstream use.
\end{enumerate}

The remainder of the paper is organized as follows.
Section~\ref{sec:benchmark} describes the ABIDES environment and ground-truth regime construction.
Section~\ref{sec:definition} formalizes crumbling quotes through deterioration steps and event clustering.
Section~\ref{sec:pipeline} presents the detection pipeline and labeling rules.
Section~\ref{sec:neural_labeling} introduces neural continuous labeling.
Section~\ref{sec:experiments} reports validation results.

\section{Market Simulator}
\label{sec:benchmark}

We implement our framework in ABIDES~\citep{byrd2019abides,byrd2020abides}, an agent-based simulator with nanosecond-resolution discrete-event scheduling, asynchronous messaging, and configurable latencies, simulating a continuous double auction for a single instrument with agent-level ground truth.
The market includes three zero-intelligence (ZI) agent types that follow exogenous signals without conditioning on limit order book (LOB) states, and two strategic agent types that adapt to LOB dynamics.
Together, these agents generate realistic order flow correlations, volatility clustering, intraday U-shaped activity, and heavy-tailed returns~\citep{vyetrenko2020get}.

\noindent\textbf{ZI Agents.}
\textit{Noise agents} submit random market orders with arrival times drawn from a $\mathrm{Beta}(1/2,1/2)$ distribution, producing the empirically observed U-shaped intraday volume~\citep{dacorogna1993geographical}, and uniform random order direction.
\textit{Value agents} observe noisy private signals of the true asset value, arriving via a Poisson process with idiosyncratic noise standard deviation $\nu > 0$, and submit limit orders buying below and selling above their valuation~\citep{kyle1985continuous}, anchoring prices to fundamentals.
\textit{Volatility agents}~\citep{xu2025learning} scale trading intensity with distance from midday, concentrating activity near market open and close, and submit larger market orders and momentum-biased limit orders, inducing volatility clustering and heavy-tailed returns~\citep{byrd2020abides}.

\noindent\textbf{Strategic Agents.}
\textit{Momentum agents} condition on recent price trends by comparing short- and long-term moving averages, generating transient price dislocations.
\textit{Adaptive market makers} follow a percent-of-volume (POV) strategy~\citep{chakraborty2011market}, placing limit orders at $L \geq 1$ price levels per side with quantities scaled by participation rate $\phi_\text{par} \in (0,1)$ and continuous cancellation and replenishment around the mid-price.
We augment this strategy with a regime-switching mechanism described below.

\noindent\textbf{Latency hierarchy.}
Agents operate under explicit latency heterogeneity: market makers occupy the lowest-latency tier reflecting co-location advantages, while background agents operate at higher latencies.
When liquidity is reduced on one side, faster background agents may consume remaining depth before quotes are restored, producing observable quote deterioration despite the market maker’s speed advantage.

\subsection{Ground-Truth Regime Construction}
\label{subsec:ground_truth}

\noindent\textbf{Skewness-based intervention.}
We induce crumbling events by introducing stochastic regime switches in the market maker’s quoting policy.
A skew parameter $\beta_t \in [\beta_{\min}, \beta_{\max}]$ controls liquidity allocation: for a fixed total quoting quantity $\tilde{Q}_t$, the ask- and bid-side quantities are
$\tilde{Q}^{\mathrm{ask}}_t = \beta_t \tilde{Q}_t$ and
$\tilde{Q}^{\mathrm{bid}}_t = (1 - \beta_t)\tilde{Q}_t$.
Symmetric quoting corresponds to $\beta_t = 0.5$, while $\beta_t > 0.5$ ($< 0.5$) induces sell-side (buy-side) bias.
At each wake-up, a Bernoulli trial with probability $\lambda$ resamples $\beta_t$ uniformly from $[\beta_{\min}, \beta_{\max}]$ and holds it fixed for duration $\Delta_{\mathrm{win}}$.
When $|\beta_t - 0.5|$ is sufficiently large, liquidity is reduced on one side, creating conditions for mechanical quote deterioration.

\noindent\textbf{Ground-truth regime variable.}
Let $Z_t^{(s)} \in \{0,1\}$ indicate whether the market maker is in crumble mode on side
$s \in \{\mathrm{bid}, \mathrm{ask}\}$:
\[
Z_t^{(\mathrm{bid})} = \mathbf{1}\{\beta_t > 0.5 + \xi\}, \qquad
Z_t^{(\mathrm{ask})} = \mathbf{1}\{\beta_t < 0.5 - \xi\},
\]
where $\xi > 0$ distinguishes meaningful skew from neutral quoting.
This variable is directly observable from the agent’s internal state and serves as ground truth for LOB-only detection evaluation.

\noindent\textbf{Ground-truth event intervals.}
Contiguous intervals during which $Z_t^{(s)} = 1$ are aggregated into event sets, with intervals separated by less than gap-merging tolerance $g^{\star} > 0$ merged.
For each side $s$, the resulting ground-truth events are
\[
\mathcal{E}^{\star}_s =
\bigl\{ [\tau^{\star}_{0,k}, \tau^{\star}_{1,k}] :
Z_t^{(s)} = 1 \ \forall t \in [\tau^{\star}_{0,k}, \tau^{\star}_{1,k}]
\bigr\},
\label{eq:ground_truth_events}
\]
where each interval represents a crumbling episode on side $s$.

\noindent\textbf{Intersection-over-union matching.}
To align predicted events $\hat{\mathcal{E}}_s$ with ground-truth events $\mathcal{E}^{\star}_s$, we use intersection-over-union (IoU).
For predicted $\hat{e} = [\hat{\tau}_0, \hat{\tau}_1]$ and ground-truth $e^{\star} = [\tau^{\star}_0, \tau^{\star}_1]$,
\[
\mathrm{IoU}(\hat{e}, e^{\star}) =
\frac{|[\hat{\tau}_0, \hat{\tau}_1] \cap [\tau^{\star}_0, \tau^{\star}_1]|}
{|[\hat{\tau}_0, \hat{\tau}_1] \cup [\tau^{\star}_0, \tau^{\star}_1]|},
\label{eq:iou}
\]
where $|\cdot|$ denotes interval length.
A predicted event $\hat{e}$ is a true positive if there exists
$e^{\star} \in \mathcal{E}^{\star}_s$ such that
$\mathrm{IoU}(\hat{e}, e^{\star}) \geq \theta_{\mathrm{IoU}}$.

\section{Crumbling Quotes Definition}
\label{sec:definition}

We define \emph{crumbling quotes} as episodes in which the best bid or ask deteriorates by multiple ticks because displayed top-of-book depth is removed faster than it is replenished, and seek to construct \emph{LOB-observable} candidate events that are mechanically consistent with visible depletion without relying on agent identities or simulator internals.
Let $\delta > 0$ denote the tick size and $\mathcal{P} = \{p_0 + k\delta : k \in \mathbb{Z}\}$ the price grid; for side $s \in \{\mathrm{bid}, \mathrm{ask}\}$ and price $p \in \mathcal{P}$, let $Q^s(p,t)$ denote the displayed depth at $(p,s)$ at time $t$.
The best bid and ask are $b(t) = \max\{p \in \mathcal{P} : Q^{\mathrm{bid}}(p,t) > 0\}, \qquad
a(t) = \min\{p \in \mathcal{P} : Q^{\mathrm{ask}}(p,t) > 0\}.$ 
For interval $(u,v]$, let $A^s(p;u,v)$, $C^s(p;u,v)$, and $E^s(p;u,v)$ denote cumulative added, canceled, and executed volume at $(p,s)$, respectively.
These satisfy the accounting identity $Q^s(p,v) - Q^s(p,u) = A^s(p;u,v) - C^s(p;u,v) - E^s(p;u,v),$ which ensures that observed depth changes must be explained by additions, cancellations, or executions at the same price level.

\noindent\textbf{Best-quote deterioration steps.}
Define the sign convention $\sigma_s = +1$ for $s = \mathrm{ask}$ and $\sigma_s = -1$ for $s = \mathrm{bid}$, and let $p_s(t)$ denote the best quote on side $s$.
Let $\{t_i\}$ be the sequence of LOB update times and write $x(t^-)$ for the left limit of any quantity $x$ at time $t$.
A \emph{deterioration step} on side $s$ at time $t_i$ occurs when
\begin{equation}
d_s(t_i) = \sigma_s \frac{p_s(t_i) - p_s(t_i^-)}{\delta} \geq 1,
\label{eq:deterioration_step}
\end{equation}
i.e., the ask increases or the bid decreases by at least one tick.
Such steps are necessary but not sufficient for crumbling, as similar movements can arise from informational repricing or message-ordering artifacts.

\noindent\textbf{Visible-depletion consistency.}
Let $\pi = p_s(t_i^-)$ denote the pre-step best quote.
Over a lookback window $(t_i - \Delta_{\mathrm{dep}}, t_i]$, where $\Delta_{\mathrm{dep}} > 0$ is the depletion horizon, we impose three conditions: $Q^s(\pi, t_i) \leq \varepsilon_Q, \qquad
C^s(\pi) + E^s(\pi) \geq (1 - \varepsilon_{\mathrm{leak}}) \, Q^s(\pi, t_i^-), \qquad
A^s(\pi) \leq \varepsilon_{\mathrm{add}} \, Q^s(\pi, t_i^-),$ where volumes are evaluated on $(t_i - \Delta_{\mathrm{dep}}, t_i]$.
Here $\varepsilon_Q \geq 0$ is the residual depth tolerance, $\varepsilon_{\mathrm{leak}} \in [0,1]$ bounds unexplained volume loss, and $\varepsilon_{\mathrm{add}} \in [0,1]$ caps replenishment relative to initial depth.
These conditions ensure that the quote moves because the displayed queue at the old best is exhausted by removals, with limited replenishment while the price deteriorates.

\noindent\textbf{Event formation.}
Crumbling episodes typically comprise multiple depletion-consistent deterioration steps occurring in close succession.
Let $\mathcal{T}_s^{\mathrm{step}}$ denote the set of times satisfying \eqref{eq:deterioration_step} and the visible-depletion conditions above.
Candidate events are formed by clustering consecutive step times separated by at most gap $g > 0$, subject to maximum duration $D_{\max} > 0$ and minimum step count $n_{\mathrm{step}} \geq 1$.
Each event is denoted $e = (\tau_0, \tau_1, s)$, spanning the first to the last step on side $s$.
Although mechanically consistent at the step level, these events may still reflect informational repricing; event-level filters and labeling are introduced in Section~\ref{sec:pipeline}.

\section{Crumbling Detection Pipeline}
\label{sec:pipeline}

Given candidate events from Section~\ref{sec:definition}, we apply event-level hard filters to filter out informational repricing, compute severity and recovery features, and define a gated binary label used for validation and as a baseline for continuous labeling (Section~\ref{sec:neural_labeling}). Throughout this section, let $\epsilon > 0$ denote a small regularization constant to prevent division by zero.

\noindent\textbf{Hard filters.}
We define four time horizons: $H_{\mathrm{pre}} > 0$ (pre-event baseline), $H_{\mathrm{post}} > 0$ (post-event observation), $H_{\mathrm{rev}} > 0$ (reversion window, with $H_{\mathrm{rev}} \leq H_{\mathrm{post}}$), and $H_{\mathrm{ref}} > 0$ (refill window).
Let $e = (\tau_0, \tau_1, s)$ be a candidate event, let $\bar{s}$ denote the opposite side, and let $\mathcal{P}_e$ be the set of price levels traversed by the best quote on side $s$ during $[\tau_0, \tau_1]$.

\noindent\textbf{Book consistency.}
Define the removed volume $V_{\mathrm{rm}}(e) = \sum_{p \in \mathcal{P}_e} ( C^s(p; \tau_0, \tau_1) + E^s(p; \tau_0, \tau_1) )$, added volume $V_{\mathrm{add}}(e) = \sum_{p \in \mathcal{P}_e} A^s(p; \tau_0, \tau_1)$, and initial depth $V_0(e) = \sum_{p \in \mathcal{P}_e} Q^s(p, \tau_0^-)$.
We require that most initial depth is accounted for by removals and that replenishment during the event is limited:
\begin{equation}
V_{\mathrm{rm}}(e) \geq (1 - \kappa_{\mathrm{miss}}) V_0(e), \qquad
\frac{V_{\mathrm{add}}(e)}{V_{\mathrm{rm}}(e) + \epsilon} \leq \kappa_{\mathrm{repr}},
\label{eq:book_consistency}
\end{equation}
where $\kappa_{\mathrm{miss}} \in [0,1]$ is the missing-volume tolerance and $\kappa_{\mathrm{repr}} \geq 0$ is the replenishment ratio bound.

\noindent\textbf{Efficient-price stability.}
Let $D_1^{\mathrm{bid}}(t) = Q^{\mathrm{bid}}(b(t), t)$ and $D_1^{\mathrm{ask}}(t) = Q^{\mathrm{ask}}(a(t), t)$ denote the displayed depth at the best bid and ask, respectively.
The microprice is defined as $p^\mu(t) = \frac{a(t) D_1^{\mathrm{bid}}(t) + b(t) D_1^{\mathrm{ask}}(t)}{D_1^{\mathrm{ask}}(t) + D_1^{\mathrm{bid}}(t)},$ and let $\hat{p}^{\mathrm{eff}}(t)$ be its exponentially smoothed version serving as an efficient-price proxy.
We require bounded displacement both during and after the event:
\begin{equation}
|\hat{p}^{\mathrm{eff}}(\tau_1^+) - \hat{p}^{\mathrm{eff}}(\tau_0^-)| \leq \kappa_{\mathrm{eff}} \delta, \qquad
|\hat{p}^{\mathrm{eff}}((\tau_1 + H_{\mathrm{post}})^+) - \hat{p}^{\mathrm{eff}}((\tau_0 - H_{\mathrm{pre}})^-)| \leq \kappa_{\mathrm{eff,post}} \delta,
\label{eq:eff_price_stability}
\end{equation}
where $\kappa_{\mathrm{eff}}$ and $\kappa_{\mathrm{eff,post}}$ are stability thresholds in tick units.

\noindent\textbf{Opposite-side stability.}
To exclude two-sided repricing, we require
$|p_{\bar{s}}(\tau_1^+) - p_{\bar{s}}(\tau_0^-)| \leq \kappa_{\mathrm{opp}} \delta$,
where $\kappa_{\mathrm{opp}}$ bounds opposite-side quote movement in ticks.

\noindent\textbf{Transience.}
Let $m(t) = (a(t) + b(t))/2$ denote the midprice.
Define the pre-event baseline $m_{\mathrm{pre}}(e) = \mathrm{median}\{m(t) : t \in [\tau_0 - H_{\mathrm{pre}}, \tau_0)\}$, the post-event level $m_{\mathrm{post}}(e) = \mathrm{median}\{m(t) : t \in (\tau_1, \tau_1 + H_{\mathrm{rev}}]\}$, and the extreme excursion $m_{\mathrm{ext}}(e) = \max_{t \in [\tau_0, \tau_1]} m(t)$ for $s = \mathrm{ask}$ (or $\min$ for $s = \mathrm{bid}$).
The reversion ratio measures how much of the price excursion persists:
\begin{equation}
\rho_{\mathrm{rev}}(e) = \frac{|m_{\mathrm{post}}(e) - m_{\mathrm{pre}}(e)|}{|m_{\mathrm{ext}}(e) - m_{\mathrm{pre}}(e)| + \epsilon} \leq \kappa_{\mathrm{rev}},
\label{eq:reversion_ratio}
\end{equation}
where $\kappa_{\mathrm{rev}} \in [0,1]$ enforces transience by requiring sufficient mean reversion.

\noindent\textbf{Event features.}
For events passing the hard filters, we compute the following features.
The \emph{walk depth} (in ticks) and \emph{depletion speed} are

$\mathrm{WD}_{\mathrm{tick}}(e) = \sigma_s \frac{p_s(\tau_1^+) - p_s(\tau_0^-)}{\delta},$ 
$\mathrm{DS}(e) = \frac{V_{\mathrm{rm}}(e)}{\tau_1 - \tau_0 + \epsilon}.$
Let $V_{\mathrm{ref}}(e) = \sum_{p \in \mathcal{P}_e} A^s(p; \tau_1, \tau_1 + H_{\mathrm{ref}})$ denote volume added during the refill window.
The \emph{refill ratio} and \emph{spread response} are $\mathrm{RR}(e) = \frac{V_{\mathrm{ref}}(e)}{V_{\mathrm{rm}}(e) + \epsilon},$ 
$\mathrm{SR}(e) = \frac{s_{\max}(e) - \mathrm{spr}_{\mathrm{pre}}(e)}{\delta},$
where $\mathrm{spr}_{\mathrm{pre}}(e)$ is the median spread over $[\tau_0 - H_{\mathrm{pre}}, \tau_0)$ and $s_{\max}(e)$ is the maximum spread during $[\tau_0, \tau_1 + H_{\mathrm{post}}]$.
The \emph{efficient-price displacement} and \emph{impact decay} are
$\mathrm{EPD}(e) = \frac{\hat{p}^{\mathrm{eff}}((\tau_1 + H_{\mathrm{post}})^+) - \hat{p}^{\mathrm{eff}}((\tau_0 - H_{\mathrm{pre}})^-)}{\delta},$ 
$\mathrm{ID}(e) = 1 - \rho_{\mathrm{rev}}(e).$

\noindent\textbf{Gated binary label.}
Let $\mathcal{F}(e) = 1$ indicate that event $e$ passes the hard filters \eqref{eq:book_consistency}--\eqref{eq:reversion_ratio}.
We define the binary crumbling label as
\begin{equation}
\ell_e = \mathbf{1}\!\left\{
\mathcal{F}(e) \wedge
\bigwedge_{f \in \{\mathrm{WD}, \mathrm{DS}, \mathrm{RR}, \mathrm{SR}, \mathrm{ID}\}} f(e) \geq \theta_f
\wedge |\mathrm{EPD}(e)| \leq \theta_{\mathrm{EPD}}
\right\},
\label{eq:labeling_rule}
\end{equation}
where $\{\theta_f\}$ are feature-specific thresholds requiring sufficient severity (walk depth, depletion speed, refill, spread widening, impact decay) while bounding efficient-price displacement to exclude informational moves.

\section{Neural Continuous Labeling via KL Calibration}
\label{sec:neural_labeling}

Section~\ref{sec:pipeline} defines mechanically interpretable candidate events and a gated binary label $\ell_e$ \ref{eq:labeling_rule}.
While the binary rule provides a baseline, it discretizes event severity and ignores interactions among features such as depletion speed, refill, spread response, and transience.
We therefore learn a continuous probabilistic crumbling label, supervised by mechanism-level regime states in simulation, to support ranking and calibration.

\noindent\textbf{Mechanics-gated probabilistic label.}
For each detected event $e = (\tau_0, \tau_1, s)$, let $x(e) \in \mathbb{R}^d$ denote the $d$-dimensional event feature vector from Section~\ref{sec:pipeline}, comprising walk depth, depletion speed, refill ratio, spread response, efficient-price displacement, and impact decay. 

\noindent\textbf{Temporal context features.}
Event-level features capture local order book dynamics but ignore cross-event dependence.
We augment $x(e)$ with three temporal features that encode market regime persistence.
Let $e_1, e_2, \ldots, e_n$ denote chronologically ordered events with end times $\tau_1^{\mathrm{end}} < \tau_2^{\mathrm{end}} < \cdots < \tau_n^{\mathrm{end}}$.
For event $e_i$ starting at $\tau_i^{\mathrm{start}}$ and lookback window $\delta_\text{rec}$, define:
\begin{align*}
\Delta t_{\mathrm{rec}}(e_i) &= \tau_i^{\mathrm{start}} - \tau_{i-1}^{\mathrm{end}}, \qquad
n_{\mathrm{rec}}(e_i; \delta) = \bigl|\{j < i : \tau_j^{\mathrm{end}} > \tau_i^{\mathrm{start}} - \delta\}\bigr|, \\[4pt]
\Sigma_{\mathrm{DS}}(e_i; \delta) &= \sum_{\substack{j < i,\; \tau_j^{\mathrm{end}} > \tau_i^{\mathrm{start}} - \delta}} \mathrm{DS}(e_j).
\end{align*}
The recovery interval $\Delta t_{\mathrm{rec}}$ measures time available for liquidity replenishment; short intervals indicate sustained stress.
The recent count $n_{\mathrm{rec}}$ captures event clustering characteristic of persistent regimes.
The cumulative depletion $\Sigma_{\mathrm{DS}}$ aggregates order flow imbalance, distinguishing isolated spikes from sustained deterioration.
The extended feature vector $\tilde{x}(e) = [x(e); \Delta t_{\mathrm{rec}}; n_{\mathrm{rec}}; \Sigma_{\mathrm{DS}}] \in \mathbb{R}^{d+3}$ replaces $x(e)$ in the encoder.
We model a latent representation $h_e$ and the corresponding score $s_e$ via:
\[
h_e = f_\phi(x(e)), \quad s_e = w^\top h_e + b,
\]
where $f_\phi : \mathbb{R}^d \to \mathbb{R}^k$ is a feedforward network with parameters $\phi$, and $(w, b) \in \mathbb{R}^k \times \mathbb{R}$ are the output layer weights and bias.
Let $\sigma(z) = 1/(1 + e^{-z})$ denote the logistic sigmoid function.
We convert the score to a probability and impose the mechanical eligibility gate from Section~\ref{sec:pipeline}:
\begin{equation}
p_\phi(e) \;\triangleq\; \mathbb{P}_\phi(Y_e = 1 \mid e) = \mathcal{F}_\eta(e) \cdot \sigma(s_e),
\label{eq:neural_prob}
\end{equation}
where $Y_e \in \{0,1\}$ indicates whether event $e$ is a true crumbling episode, and
$\mathcal{F}_\eta(e) \in \{0,1\}$ indicates whether $e$ satisfies the hard filters
\eqref{eq:book_consistency}--\eqref{eq:reversion_ratio}.
The parameter set $\eta$ denotes the associated threshold hyperparameters
(e.g., $\kappa_{\mathrm{eff}}$, $\kappa_{\mathrm{opp}}$, $\kappa_{\mathrm{rev}}$).
Equation~\ref{eq:neural_prob} can be interpreted as a multiplicative gating layer:
$\mathcal{F}_\eta(e)$ may be fixed, as in this work, or learned jointly with the classifier
via a differentiable relaxation of the hard indicators.
We fix $\eta$ here because the simulated environment yields clean regime-labeled events;
in empirical settings, calibrating these necessary conditions may be required and is
supported by the framework.

\noindent\textbf{Supervision from ABIDES regime states.}
The simulator provides side-specific regime indicators $Z_t^{(s)}$ (Section~\ref{sec:benchmark}) and ground-truth interval sets $\mathcal{E}^\star_s$.
For each detected event $e = (\tau_0, \tau_1, s)$, we define a binary target $y_e \in \{0, 1\}$ by temporal overlap with ground truth on the same side:
$y_e = 1$ if there exists $e^\star \in \mathcal{E}^\star_s$ such that $\mathrm{IoU}(e, e^\star) \geq \theta_{\mathrm{IoU}}$, and $y_e = 0$ otherwise, where $\theta_{\mathrm{IoU}} \in (0, 1]$ is the matching threshold defined in Section~\ref{sec:benchmark}.
This supervision depends only on agent regime states and event timing, not on the feature thresholds in \ref{eq:labeling_rule}.
A continuous target $y_e \in [0, 1]$ can alternatively be defined using the maximum overlap fraction across ground-truth events.

\noindent\textbf{KL calibration objective.}
Let $\mathcal{D}$ denote the dataset of detected candidate events, and let $\phi$ denote the parameters of the neural labeling model.
Interpreting $p_\phi(e)$ as the success probability of a Bernoulli random variable $Y_e$, we train the model by minimizing the KL divergence
$\mathrm{KL}\!\left(\mathrm{Bern}(y_e)\,\|\,\mathrm{Bern}(p_\phi(e))\right)$
between the empirical labels and the model predictions.
This objective is equivalent to the binary cross-entropy loss: $\mathcal{L}_{\mathrm{KL}}(\phi)
= -\sum_{e \in \mathcal{D}} \bigl[
y_e \log p_\phi(e) + (1 - y_e)\log(1 - p_\phi(e))
\bigr].$
\section{Experiments}
\label{sec:experiments}
\noindent\textbf{Simulation Configuration.}
We simulate five consecutive equity trading sessions from 9:30 AM to 4:00 PM ET with tick size $\delta = 1$ cent.
The market consists of heterogeneous agents commonly used in limit order book models: 150 noise agents, 100 value agents with idiosyncratic noise $\nu = 10$ cents, 5 momentum agents, and one adaptive market maker posting at $L = 10$ price levels per side with participation rate $\phi_\text{par} = 0.025$.
This agent mix generates order flow from noise trading, fundamental valuation, and short-term trend-following.
Message latencies are tiered by agent type, with the market maker operating at 100~$\mu$s and background agents (noise, value, momentum) operating at 1~ms, reflecting latency heterogeneity in electronic markets.
The market maker’s quote-skewness parameter $\beta_t \in [0.1, 0.9]$ follows stochastic regime switches triggered by Bernoulli trials with probability $\lambda = 0.05$ and persisting for $\Delta_{\mathrm{win}} = 1$ second, producing finite-duration, side-specific liquidity withdrawal.
The ground-truth indicator $Z_t^{(s)}$ uses threshold $\xi = 0.15$ to identify deviations from symmetric quoting.
We evaluate robustness across four regimes: baseline, bull with upward drift ($+\$0.10$/hour), bear with downward drift ($-\$0.10$/hour), and high-volatility with 50 additional volatility agents.

\noindent\textbf{Detection Parameters.}
\textit{Candidate detection:} depletion lookback $\Delta_{\mathrm{dep}} = 100$ ms; queue exhaustion $\varepsilon_Q = 0.05 \cdot Q_{\mathrm{pre}}$; removal tolerance $\varepsilon_{\mathrm{leak}} = 0.10$; replenishment cap $\varepsilon_{\mathrm{add}} = 0.15$; clustering gap $g = 200$ ms; max duration $D_{\max} = 2$ s; min steps $n_{\mathrm{step}} = 4$.
\textit{Hard filters:} time horizons $H_{\mathrm{pre}} = H_{\mathrm{post}} = H_{\mathrm{ref}} = 1$ s, $H_{\mathrm{rev}} = 3$ s; book consistency $\kappa_{\mathrm{miss}} = 0.05$, $\kappa_{\mathrm{repr}} = 0.20$; efficient-price stability $\kappa_{\mathrm{eff}} = 5$ ticks, $\kappa_{\mathrm{eff,post}} = 8$ ticks; one-sidedness $\kappa_{\mathrm{opp}} = 5$ ticks; transience $\kappa_{\mathrm{rev}} = 0.6$.
\textit{Labeling thresholds:} $\theta_{\mathrm{WD}} = 2$ ticks, $\theta_{\mathrm{SR}} = 1$ tick, $\theta_{\mathrm{EPD}} = 6$ ticks, $\theta_{\mathrm{ID}} = 0.3$; $\theta_{\mathrm{DS}}$ and $\theta_{\mathrm{RR}}$ calibrated at 5th percentile.
\textit{Evaluation:} IoU matching with $\theta_{\mathrm{IoU}} = 0.3$.

\begin{figure}[ht]
    \centering
    \begin{subfigure}[b]{0.24\columnwidth}
        \centering
        \includegraphics[width=\linewidth]{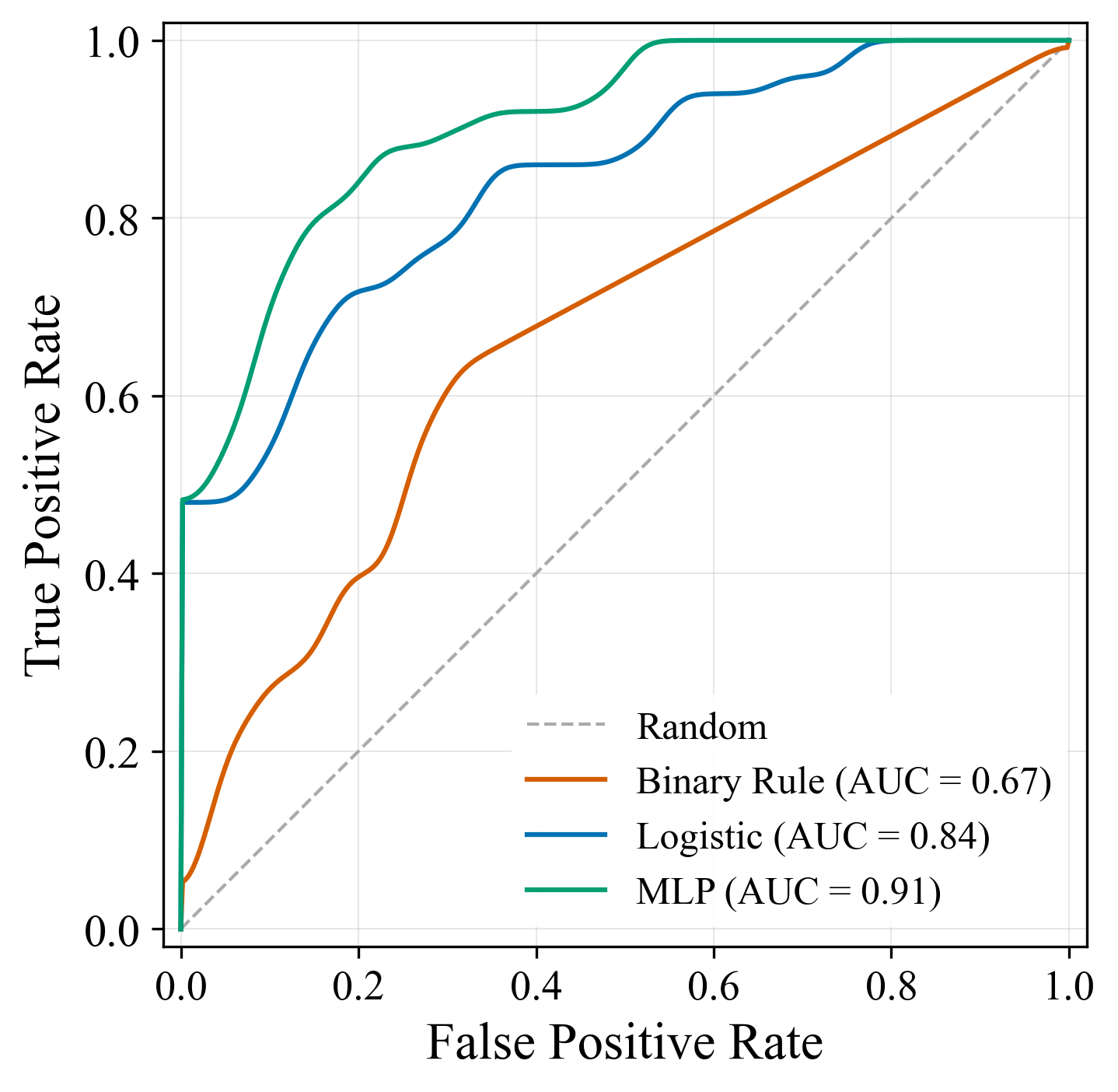}
        \caption{Baseline}
    \end{subfigure}
    \hfill
    \begin{subfigure}[b]{0.24\columnwidth}
        \centering
        \includegraphics[width=\linewidth]{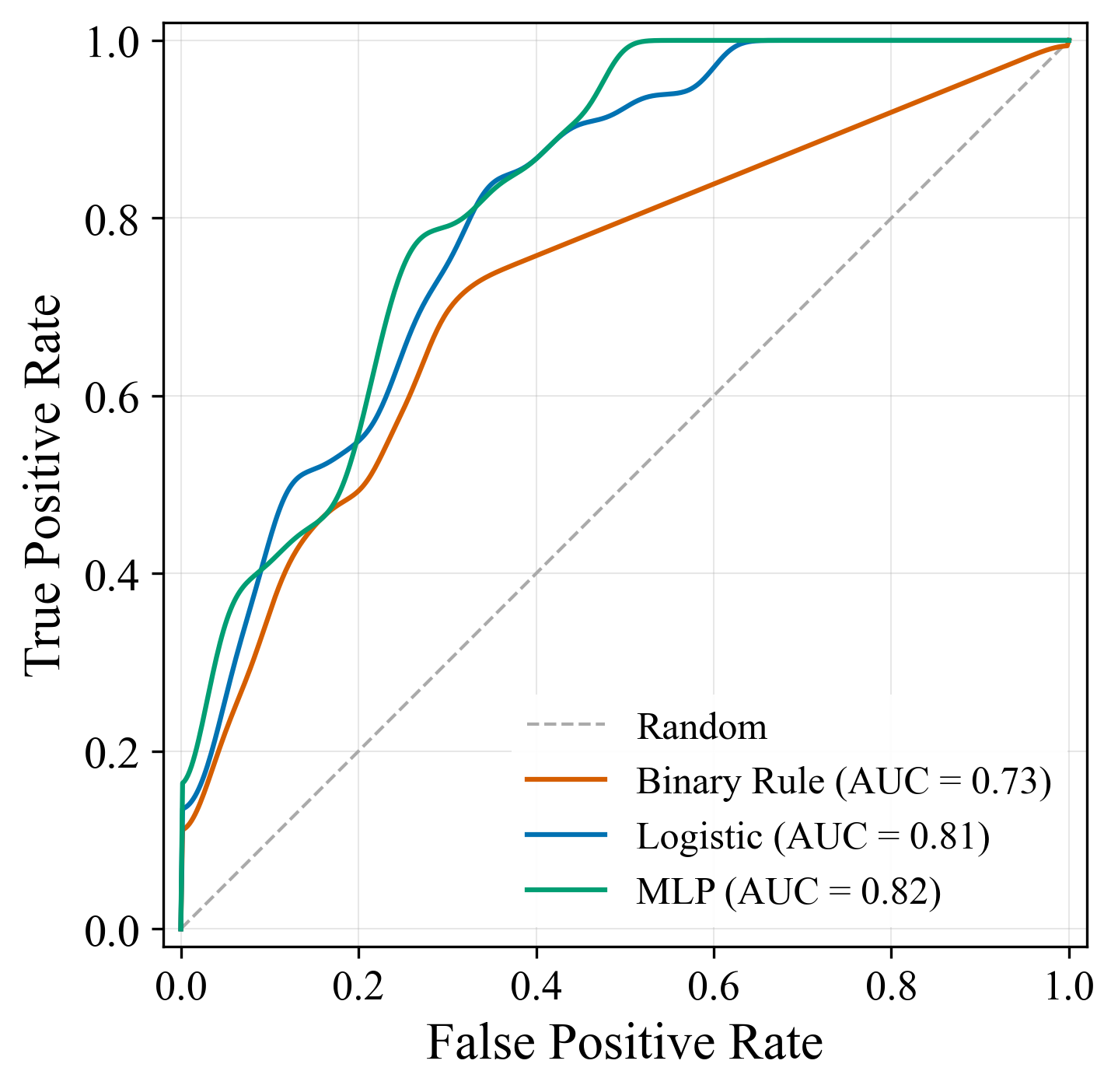}
        \caption{Bull Market}
    \end{subfigure}
    \hfill
    \begin{subfigure}[b]{0.24\columnwidth}
        \centering
        \includegraphics[width=\linewidth]{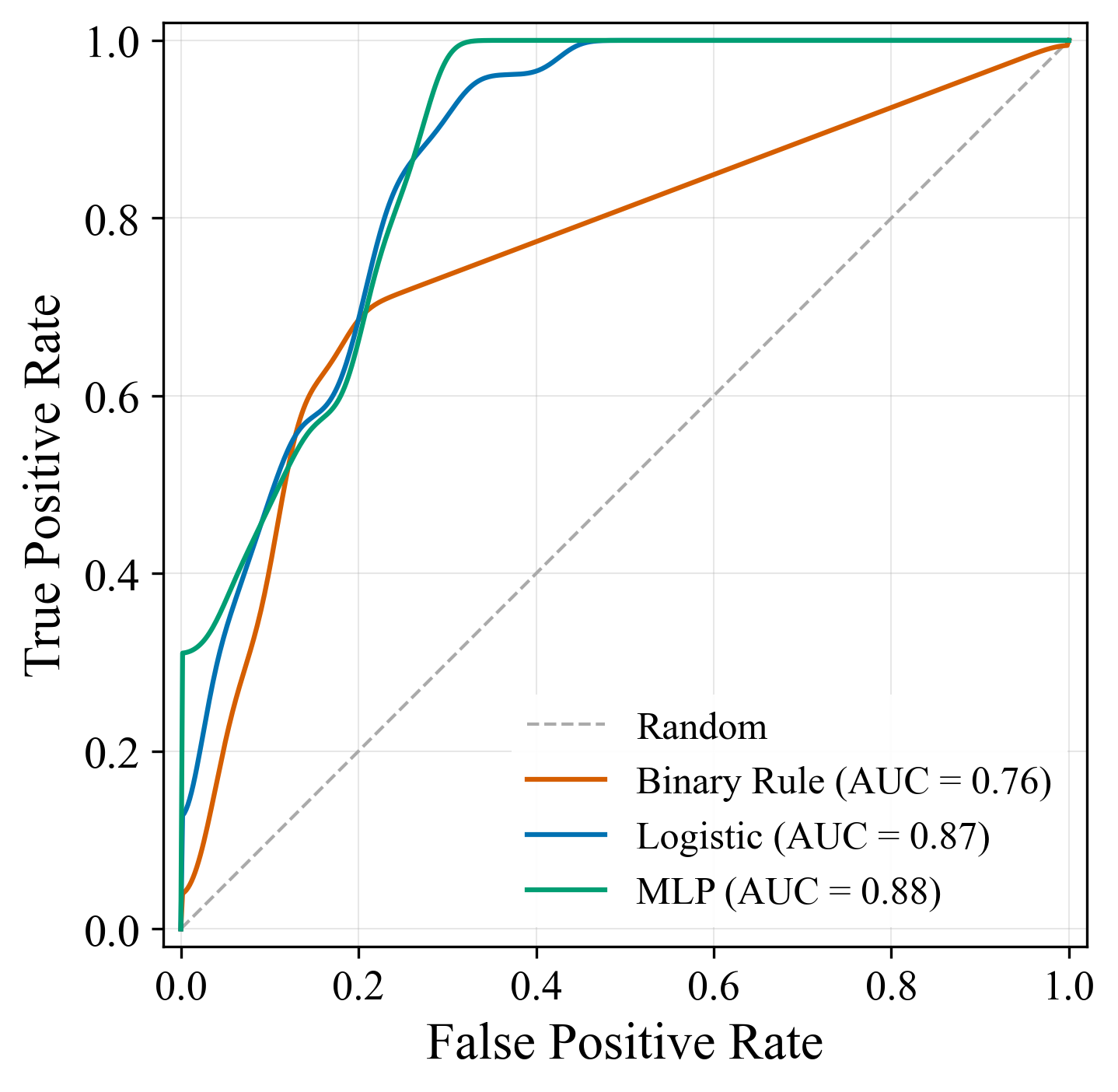}
        \caption{Bear Market}
    \end{subfigure}
    \hfill
    \begin{subfigure}[b]{0.24\columnwidth}
        \centering
        \includegraphics[width=\linewidth]{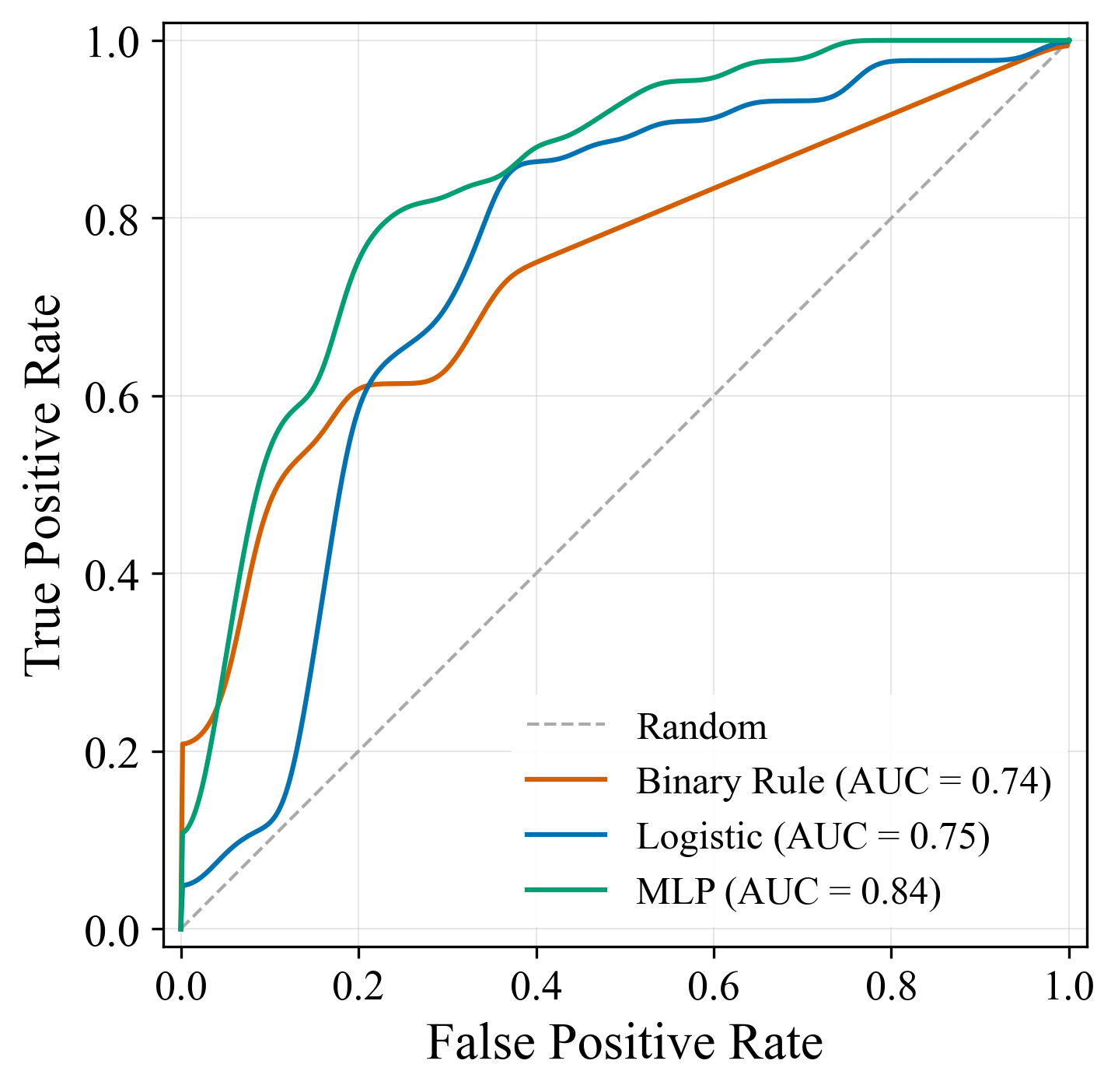}
        \caption{High Volatility}
    \end{subfigure}

    \caption{ROC curves for crumbling detection across market regimes. Each panel compares the Binary Rule (orange), Logistic Regression (blue), and MLP (green) classifiers. (a) Baseline regime with standard agent population. (b) Bull market with upward price drift. (c) Bear market with downward price drift. (d) High-volatility regime with momentum-biased order flow.}
    
    \label{fig:regime_roc}
\end{figure}

\noindent\textbf{Neural Continuous Labeling.} The encoder $f_\phi$ is a three-layer Multi-Layer Perceptron (MLP) with hidden dimensions [64, 32], LayerNorm, GELU activation, and dropout rate 0.1.
Features are normalized using robust scaling (median and interquartile range).
We train with AdamW optimizer using learning rate $5 \times 10^{-3}$, weight decay $0.05$, and batch size 32.
We use a 70/15/15 train/validation/test split, and also evaluate logistic regression with RBF kernel approximation as a baseline to isolate the contribution of nonlinear feature interactions.

\subsection{Results}

\noindent\textbf{Discrimination Performance.}
Figure~\ref{fig:regime_roc} compares ROC curves across market regimes.
In the baseline setting (panel~a), the Binary Rule attains AUC~=~0.67, Logistic Regression improves to 0.84, and the continuous labeling model achieves the highest discrimination with AUC~=~0.91.
This ordering persists across stressed regimes (panels~b--d), although the High Volatility condition reduces the gap between learned models and the Binary Rule, consistent with noisier microstructure signals.
Overall, the results indicate that the continuous labeling model captures robust crumbling dynamics.

\noindent\textbf{Ablation Study: Temporal Dependence Structure.}
A key concern in simulation-calibrated labeling is sensitivity to the temporal mechanism generating regime switches.
We therefore replace the baseline memoryless switching of the market maker's crumbling regime with a self-exciting Hawkes process, inducing clustered crumbling intervals.
Event times follow intensity
$\lambda(t) = \lambda_0 + \sum_{t_i < t} \gamma \exp(-\omega(t - t_i))$
with $\lambda_0 = 0.03$, $\gamma = 0.20$, and $\omega = 0.08$, producing strongly clustered arrivals.
\begin{figure}[ht]
\begin{center}
\includegraphics[width=\textwidth]{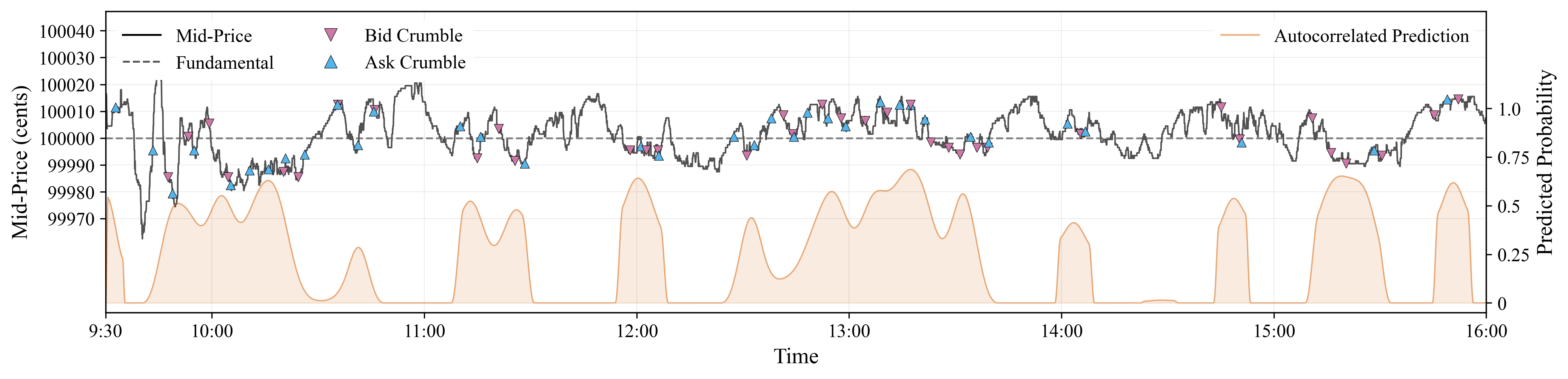}
\end{center}
\caption{Mid-price trajectory (top) and event-level crumbling outputs over one trading day under Hawkes-driven regime switches.
Triangles mark binary detections (down: bid-side; up: ask-side).
The lower panel shows continuous crumbling probabilities from the neural labeling model, exhibiting persistent high-probability segments separated by quiet intervals.}
\label{fig:price_dynamics}
\end{figure}
Figure~\ref{fig:price_dynamics} presents continuous crumbling probabilities from the neural labeling model under Hawkes-driven regime dynamics.
Elevated probability regions exhibit strong temporal correspondence with ground-truth detections, demonstrating that the learned representation captures the clustering structure of the self-exciting arrival process.
High-probability intervals coincide with periods of mechanical liquidity withdrawal, while near-zero outputs indicate normal market conditions, validating the discriminative capacity of the framework.
\begin{figure}[ht]
\begin{minipage}[ht]{0.4\textwidth}
\vspace{0pt}
\centering
\includegraphics[width=0.7\linewidth]{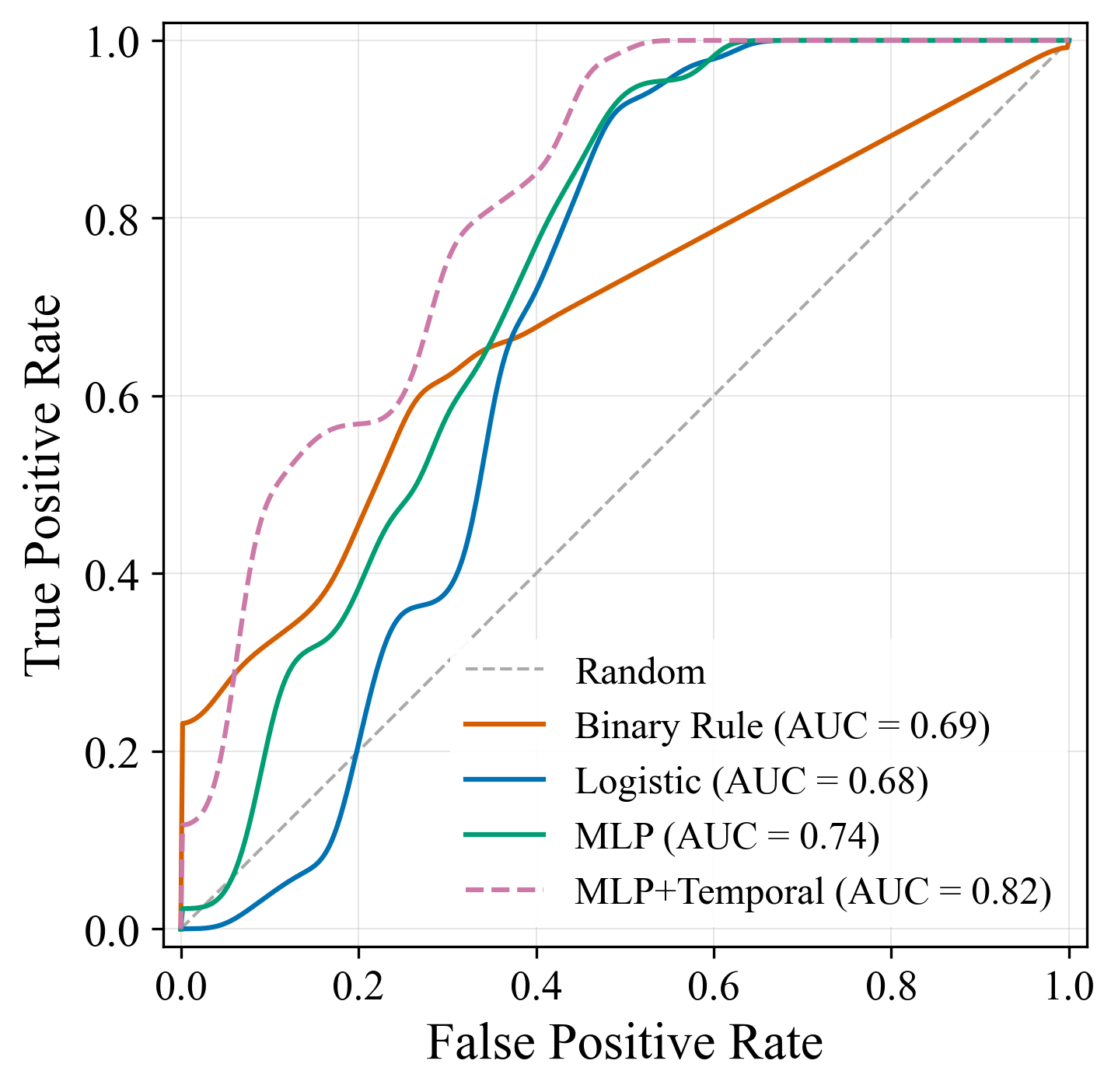}
\captionof{figure}{ROC curves for crumbling detection. Adding temporal context features ($\Delta t_{\mathrm{rec}}$, $n_{\mathrm{rec}}$, $\Sigma_{\mathrm{DS}}$) to the MLP yields the highest discrimination, outperforming the binary rule baseline and logistic regression.}
\label{fig:roc_temporal}
\end{minipage}
\hfill
\begin{minipage}[ht]{0.53\textwidth}
\vspace{0pt}
\noindent\textbf{Temporal Feature Contribution.}
Figure~\ref{fig:roc_temporal} compares detection methods with and without temporal context.
We set the lookback window $\delta_\text{rec} = 60$s, matching typical market maker inventory adjustment horizons.
Short recovery intervals $\Delta t_{\mathrm{rec}}$ indicate sustained liquidity stress where the market has insufficient time to replenish depth between deterioration events.
High recent counts $n_{\mathrm{rec}}$ signal regime persistence characteristic of informed trading pressure rather than transient noise.
Elevated cumulative depletion $\Sigma_{\mathrm{DS}}$ reflects persistent order flow imbalance.
Together, these features enable the classifier to distinguish isolated liquidity events from clustered episodes arising under persistent crumbling regimes.
\end{minipage}
\end{figure}

\bibliography{iclr2026_conference}
\bibliographystyle{iclr2026_conference}

\end{document}